\documentclass[11pt,table,reqno]{amsart}

\usepackage{amsfonts,amssymb,mathrsfs}
\usepackage{times}
\usepackage{array}
\usepackage{graphicx}
\usepackage[all]{xy}
\usepackage{amscd}
\usepackage{color}
\usepackage{slashbox}
\usepackage{colortbl}
\usepackage{pifont}
\usepackage{amsmath}
\usepackage[onehalfspacing]{setspace}
\usepackage{pdflscape}
\usepackage{colortbl}
\usepackage[table]{xcolor}
\usepackage{tabularx,booktabs}
\usepackage{boxedminipage}
\usepackage{graphicx}
\usepackage{rotating}
\usepackage{booktabs}
\usepackage{longtable}
\usepackage{subfigure}
\usepackage{wrapfig}
\usepackage{lipsum}
\usepackage{multirow}
\usepackage{color}
\usepackage[table]{xcolor}
\usepackage{booktabs}
\usepackage{ctable}
\usepackage{array}
\usepackage[latin1]{inputenc}
\usepackage{tikz}
\usepackage{tcolorbox}
\usepackage{blindtext}
\usepackage{graphicx}

\newcolumntype{U}{>{\columncolor[gray]{0.8}}c}
\newcolumntype{U}{>{\columncolor[red]{0.8}}c}
\newcolumntype{U}{>{\color[red]{0.8}}c}
\newcolumntype{U}{>{\xcolor[red]{0.8}}c}
\newcolumntype{U}{>{\colorbtl[red]{0.8}}c}
\numberwithin{equation}{section}
\newcolumntype{P}[1]{>{\centering\arraybackslash}p{#1}}
\newcolumntype{M}[1]{>{\centering\arraybackslash}m{#1}}

\newtheorem{thm}{Theorem}[section]

\newtheorem{cor}[thm]{Corollary}

\newtheorem{prop}[thm]{Proposition}

\theoremstyle{definition}
\newtheorem{Definition}[thm]{Definition}

\newtheorem{rmk}[thm]{Remark}

\begin{document}
\baselineskip 15.6225pt

\title{Digital (co)homology modules and digital Pontryagin algebras}

\thanks{
This work was supported by the National Research Foundation of Korea (NRF) grant funded by the Korean government (MSIT) (No. 2018R1A2B6004407).}
\thanks{~~Tel.: +82-63-270-3367.}

\author{Dae-Woong Lee}

\address{Department of Mathematics, and Institute of Pure and Applied Mathematics, Jeonbuk National University,
567 Baekje-daero, Deokjin-gu, Jeonju-si, Jeollabuk-do 54896,
Republic of Korea}
\email{dwlee@jbnu.ac.kr}

\subjclass[2010]{Primary 16T05; Secondary 16T15, 68U10, 57T25, 55N35.}
\keywords{Digital homotopy, digital $n$-simplex, digital (co)homology module, digital primitive (co)homology class, digital convolution, digital Pontryagin algebra.}

%\renewcommand\leftmark{\centerline{Dae-Woong Lee}}
%\renewcommand\rightmark{\centerline{   }}

% Abstract (Do not insert blank lines, i.e. \\)
\begin{abstract}
In the current study, we explore digital homology and cohomology modules, and investigate their fundamental properties on pointed digital images. We also examine pointed digital Hopf spaces and base point preserving digital Hopf functions between the pointed digital Hopf spaces with suitable digital multiplications, and explore the digital primitive homology and cohomology classes, the digital Pontryagin algebras and coalgebras on the digital Hopf spaces as digital images.
\end{abstract}

\maketitle

\section{Introduction}

\subsection{History and Hopf space}
Digital geometry deals with bounded and finite discrete sets in the sense of classical topology, which is considered to be digital images or digitalized models of bounded and finite subsets of the lattice points in the Euclidean space.
The homology and cohomology modules, higher homotopy groups, stable homotopy groups and equivariant homotopy groups are useful algebraic and topological tools to solve a large number of problems of algebraic geometry and algebraic topology.
%and are in practice the fundamental algebraic tools.
In the same lode, the digital counterparts of classical homology and cohomology modules can be important
gadgets to classify (pointed) digital images from the point of view for the digital version of the homotopy type, mathematical morphology, and image synthesis. In particular, the informal definitions of many terms in elementary homotopy and simplicial homology theory based on a digital picture on $\mathbb{Z}^2$ or $\mathbb{Z}^3$ were nicely presented in \cite{MO2}, \cite{MO1}, \cite{AKO}, \cite{BKO} and \cite{EK2}; see also \cite{L7} for digital quasi co-Hopf space.

In the 20th century, a lot of interesting and remarkable results on Lie groups and (pointed) Hopf spaces, as the Eckmann-Hilton dual notions of (pointed) co-Hopf spaces, have been widely investigated and developed by suitable methods for CW-spaces and usual topological spaces.
The (pointed) Hopf spaces has been the direct outgrowth of compact Lie groups in classical homotopy theory, as described in \cite{J}, \cite{MC} and \cite{ML3}. Indeed, a pointed Hopf space is a triple $(Y,y_0, m_Y )$ which consists of a pointed topological space $(Y,y_0)$ and a base point preserving continuous multiplication $m_Y : Y \times Y \rightarrow Y$ such that a constant function $e_{y_0} : Y \rightarrow Y$ at $y_0$ plays a role of a homotopy identity, i.e.,
$m_Y(c_{y_0}, y) = y = m_Y(y, c_{y_0})$ for all $y \in Y$ in the pointed homotopy category. The notion of (pointed) Hopf spaces is one of the Eckmann-Hilton dual notions of a co-Hopf space; see \cite{G}, \cite{L1}, \cite{L0}, \cite{MLD}, \cite{L2}, \cite{ML}, \cite{L3}, \cite{L4}, \cite{L5}, \cite{L6}, \cite{L10}, \cite{LS} and \cite{L9} for topics related to those basic notions. It can be seen that all (pointed) Lie groups are (pointed) Hopf spaces. In general, the Hopf spaces lack associative and inversive properties and do not have the structure of usual topological manifolds at all.

The multiplication in a Hopf space provides the homology modules of a Hopf space with an algebraic structure which is natural with respect to Hopf functions. In fact, we can construct the algebra structure at homology level which is called as the {\it Pontryagin algebra} of the Hopf space. Under suitable conditions, the diagonal function gives the non-negatively graded homology module a coalgebra structure. The two algebraic structures are related to each other and covert the non-negatively graded homology modules into a classical Hopf algebra.

\subsection{Motivation}
There are a few standard approaches for considering a digital analogue of the well-known usual topology on $\mathbb{R}^n$ such as the graph-theoretic approach, the imbedding approach, and the axiomatic approach \cite{EL}. From this point of view, we need to investigate another approach to study digital topology out of classical algebraic topology. In the present paper, we introduce another consideration of a digital analogue as the so-called algebraic approach from the classical homology, cohomology and Pontryagin algebra. More precisely, the current study is concerned with setting up more algebraic invariants and their fundamental properties of digital homology and cohomology modules over a commutative ring with identity for digital image with an adjacent relation which are based on the classical
homology and cohomology groups of topological spaces in algebraic topology.

\subsection{Organization of the paper}
The current paper has been organized as follows.
%We now give a short overview of the structure of this paper.
%The paper is organized as follows:
In Section \ref{G} we introduce the general notions of digital images with $k_X$-adjacent
relations. In Section \ref{H} we define a digital $n$-simplex, the digital $n$-chains, and the digital homology and cohomology modules over a commutative ring $R$ with identity of digital images. We would then construct a covariant and contravariant functor from a category of
digital images and  digital continuous functions to the one of
$R$-modules and $R$-module homomorphisms. We also investigate some
fundamental and interesting properties of digital homology and cohomology modules and the primitive homology classes of digital images.
Moreover, we would show that the digital version of the dimension axiom is guaranteed as one of the Eilenberg-Steenrod axioms in algebraic topology. In Section \ref{R} we consider a pointed digital Hopf space together with digital multiplications, digital homotopy associative and commutative multiplications, and base point preserving digital Hopf functions between the pointed digital Hopf spaces with digital multiplications based on the pointed digital sets. We also explore the important properties of digital primitive cohomology classes and digital Pontryagin algebras in digital Hopf spaces as digital images.

\bigskip

\section{Preliminaries}\label{G}

Let $\mathbb{Z}$ be the ring of integers and $\mathbb{R}$ the field of all real
numbers. Let $\mathbb{Z}^n$ be the set of all lattice points
in the  $n$-dimensional Euclidean space $\mathbb{R}^n$. A {\it digital image} is a pair $(X, k_X)$, where $X$ is a bounded and finite subset of $\mathbb{Z}^n \subsetneq \mathbb{R}^n$ and $k_X$ indicates some adjacent relation between the members of $X$; see below.

For an integer $u$ with $1 \leq u \leq n$, we will first define an adjacent relation
of a digital image in $\mathbb{Z}^n$ as follows.

\begin{Definition} {\rm (\cite{LB3, Han})
Two points $p =(p_1,p_2,\ldots, p_n)$ and $q =(q_1,q_2,\ldots,q_n)$ with $p \neq q$ in $\mathbb{Z}^n$
are {\it $k(u,n)$-adjacent} if
\begin{enumerate}
\item[\rm{(1)}] there are at most $u$ distinct indices $i$ with the property $\vert p_i - q_i\vert =1$; and
\item[\rm{(2)}] if $\vert p_j - q_j \vert \neq 1$, then $p_j = q_j$ for all indices $j$.
\end{enumerate}
}\end{Definition}

A $k(u,n)$-adjacency relation on $\mathbb{Z}^n$ may be denoted by the
number of points that are $k(u,n)$-adjacent to a point $p \in \mathbb{Z}^n$. Moreover,
\begin{itemize}
\item the $k(1,1)$-adjacent points of $\mathbb{Z}$ are called $2$-adjacent;
\item the $k(1,2)$-adjacent points of $\mathbb{Z}^2$ are called $4$-adjacent, and the $k(2,2)$-adjacent points in $\mathbb{Z}^2$ are called $8$-adjacent;
\item the $k(1,3)$-adjacent points of $\mathbb{Z}^3$ are called $6$-adjacent, the $k(2,3)$-adjacent points of $\mathbb{Z}^3$ are called $18$-adjacent, and the $k(3,3)$-adjacent points of $\mathbb{Z}^3$ are called $26$-adjacent;
\item the $k(1,4)$-, $k(2,4)$-, $k(3,4)$-, and $k(4,4)$-adjacent points of $\mathbb{Z}^4$ are called $8$-adjacent, $32$-adjacent, $64$-adjacent, and $80$-adjacent, respectively.
\end{itemize}
We note that the above number $k(u,n), 1 \leq u \leq n$ is just the
cardinality of the set of lattice points which have the
$k(u,n)$-adjacent relations centered at $p$ in $\mathbb{Z}^n$. We
mostly denote $k(u,n)$-adjacent relation on a digital image $X$ by $k_X$-adjacent relation for short
if there is no chance of ambiguity.

\begin{Definition} {\rm (\cite {LB0, LB1})  A digital image $(X,k_X)$ in $\mathbb{Z}^n$ is said to be {\it $k_X$-connected} if
for every pair of points $\{x,y\} \subset X$ with $x \neq y$, there
exists a set $P= \{ x_0 ,x_1, \ldots, x_s \} \subset X$ of $s+1$
distinct points such that $x=x_0, x_s =y$, and $x_i$ and $x_{i+1}$
are $k_X$-adjacent for $i = 0,1,\ldots,s-1$.
%The {\it length} of the set $P$ is the number $s$.
}\end{Definition}

The following is a minor modification of an earlier definition of a
digital continuous function given in \cite[Definition 2.3]{LB1}; see also
\cite{R}.

\begin{Definition} {\rm \label{DCdef} Let $(X,k_X)$ and $(Y,k_Y)$ be the digital
images with $k_X$-adjacent and $k_Y$-adjacent relations,
respectively. A function $f : X \rightarrow Y$ from $(X,k_X)$ to $(Y,k_Y)$ is said to be a {\it
$(k_X,k_Y)$-continuous function} if the image of every
$k_X$-connected subset of the digital image $X$ under $f$ is a $k_Y$-connected
subset of $Y$.
}\end{Definition}

Let
$(X,k_X)$, $(Y,k_Y)$ and $(Z,k_Z)$ be the digital images. If
$f: X \rightarrow Y$ is a $(k_X,k_Y)$-continuous function and $g: Y \rightarrow Z$ is a $(k_Y,k_Z)$-continuous function, then it is not difficult to show that the composite $g \circ f : X \rightarrow Z$ of $f$ and $g$ is $(k_X,k_Z)$-continuous. Thus, it is possible to construct the category $\mathcal D$ of digital images and digital continuous functions; that is, the object classes of $\mathcal D$ are digital images and the morphism classes are digital continuous functions.

\begin{Definition} {\rm (\cite {LB0, LB2}) Let $a,b \in \mathbb{Z}$ with $a < b$. A {\it digital interval} is a set of the form
$
[a,b]_{\mathbb{Z}} = \{ z \in \mathbb{Z} ~\vert~ a \leq z \leq b \}
$
in which 2-adjacent relation in $\mathbb{Z}$ is assumed.
}\end{Definition}

\begin{Definition} {\rm (\cite {K, LB1, LB3}) Let $(X,k_X)$ and $(Y,k_Y)$ be digital images with $k_X$-adjacent and $k_Y$-adjacent relations, respectively, and let $f,g : X \rightarrow Y$ be $(k_X, k_Y)$-continuous functions. Suppose that there is a positive integer $m$ and a function $F : X \times [0,m]_{\mathbb{Z}} \rightarrow Y$ such that
\begin{itemize}
\item $F(x,0) = f(x)$ and $F(x,m) =g(x)$ for all $x \in X$;
\item the induced function $F_x : [0,m]_{\mathbb{Z}} \rightarrow Y, x \in X$ defined by
 $F_x (t) = F(x,t)$ for all $t \in [0,m]_{\mathbb{Z}}$ is $(2, k_Y)$-continuous; and
\item the induced function $F_t : X \rightarrow Y, t \in [0, m]_{\mathbb{Z}}$ defined by $F_t (x) = F(x,t)$ for all $x \in X$ is $(k_X, k_Y)$-continuous.
\end{itemize}
Then, $F$ is called a {\it digital $(k_X,k_Y)$-homotopy} between $f$
and $g$, written as $F: f \simeq_{(k_X,k_Y)} g$, and $f$ and $g$ are
called {\it digitally  $(k_X,k_Y)$-homotopic} in $Y$.
}\end{Definition}

We end this section giving the description of the pointed versions of digital images to develop the pointed digital category as follows.

\begin{Definition} {\rm (\cite{L8, L11}) A {\it pointed digital image} with $k_X$-adjacent relation is a triple $(X,x_0, k_X)$, where $X$ is a digital image and $x_0 \in X$. In this case, $x_0$ is said to be a {\it base point}
of $(X,x_0, k_X)$. A {\it pointed digital continuous function}
$$
f : (X,x_0, k_X) \rightarrow (Y,y_0, k_Y)
$$
is a $(k_X,k_Y)$-continuous function from $(X,x_0)$ to $(Y,y_0)$ such that
$$
f(x_0) = y_0.
$$
A digital homotopy
$$
F : X \times [0,m]_{\mathbb{Z}} \rightarrow Y
$$
between pointed digital continuous functions $f$ and $g$ is said to
be {\it pointed digital $(k_X,k_Y)$-homotopy} between $f$ and $g$ if $F(x_0,t) = y_0$ for all $t
\in [0, m]_{\mathbb{Z}}$.
}\end{Definition}

We now construct the so-called pointed digital category $\mathcal D_*$ of pointed digital images and base point preserving digital continuous functions; that is, the object classes of $\mathcal D_*$ are pointed digital images and the morphism classes are base point preserving digital continuous functions.

\bigskip

\section{Digital homology and cohomology modules}\label{H}

In this section, we consider the digital homology and cohomology modules over a commutative ring $R$ with identity `$1_R$' (compare with \cite{BKO} in the case of digital simplicial homology groups).
For $i = 0,1,\ldots,n$, we let $e_i$ be the point in $\mathbb{Z}^{n+1}$ having
coordinates all zeros except for 1 in the $(1+i)$th coordinate; that is,
$e_0 =(1,0,0,\ldots,0)$, $e_1 = (0,1,0,\ldots, 0)$, $\ldots$, and $e_n = (0,0,\ldots, 0, 1)$ in $\mathbb{Z}^{n+1}$.

\begin{Definition} {\rm  A {\it digital convex combination} of points $e_0,e_1,\ldots,e_n$ in $\mathbb{Z}^{n+1}$ is a point $x$ with
$$
x = r_0 e_0 + r_1 e_1 + \cdots + r_n e_n ,
$$
where $\sum_{i=0}^n r_i =1$, and $r_i = 0$ or $1$. The coefficients
$(r_0, r_1, \ldots, r_n)$ of
$$
x = r_0 e_0 + r_1 e_1 + \cdots + r_n e_n
$$
are called the {\it digital barycentric coordinates} of $x$.
}\end{Definition}

Unlike the classical convex combination of points,
it can be easily verified that $x$ is a digital convex combination of $e_0,e_1,\ldots,e_n$ if and only if $x$ is an element of
$\{ e_0,e_1,\ldots,e_n\}$.
We let $\Delta^n$ be the set of all digital convex combinations
of points $e_0,e_1,\ldots,e_n$ in $\mathbb{Z}^{n+1}$; that is, $\Delta^n = \{ e_0,e_1,\ldots,e_n \}$, which is completely different from the usual convex combinations in algebraic topology when $n \geq 1$. Considering $\Delta^n$ as the digital image with $k(2,n+1)$-adjacent relation, we can see that it is
$k(2,n+1)$-connected. We denote the $k(2,n+1)$-adjacent relation in the digital image $\Delta^n$ by $k_{\Delta^n}$ for our notational convenience, as mentioned earlier.

\begin{Definition} {\rm
Using the $k_{\Delta^n}$-adjacent relation in the digital image $\Delta^n = \{ e_0,e_1,\ldots,e_n \}$, we consider an {\it orientation} of $\Delta^n$ as a linear ordering of its vertices, and call it a {\it digital standard $n$-simplex} with linear order.
}\end{Definition}

\begin{Definition} {\rm Let $(X,k_X)$ be a digital image with $k_X$-adjacent relation. A {\it digital $n$-simplex} in $(X,k_X)$ is a $(k_{\Delta^n},k_X)$-continuous function
$
\sigma : (\Delta^n, k_{\Delta^n}) \rightarrow (X,k_X) ,
$
where $\Delta^n$ is the digital standard $n$-simplex.
}\end{Definition}

A constant function is an example of a digital $n$-simplex in $(X,k_X)$. Moreover,
for every $\{e_i , e_j\} \subset \Delta^n$ so that $e_i$ and $e_j$
are $k_{\Delta^n}$-adjacent in $\Delta^n$, either $\sigma (e_i ) =
\sigma (e_j )$, or $\sigma (e_i )$ and $\sigma (e_j )$ are
$k_X$-adjacent in $(X,k_X)$.

\begin{Definition} {\rm Let $R$ be a commutative ring with identity `$1_R$' and let $(X,k_X)$ be a digital image with $k_X$-adjacent relation. For each $n \geq 0$,
we define $dC_n(X;R)$ to be the non-negatively graded free $R$-module with basis all digital
$n$-simplexes in $(X,k_X)$. The elements of $dC_n(X;R)$ are called {\it digital $n$-chains}
in $(X,k_X)$.
}\end{Definition}

We note that the oriented boundary of a digital $n$-simplex $\sigma :
(\Delta^n, k_{\Delta^n})  \rightarrow (X,k_X)$ has to be $\sum_{i=0}^n (-1)^i (\sigma
\vert_{\{ e_0,\ldots,\hat e_i, \ldots, e_n \}})$, where the symbol
$\hat e_i$ means that the vertex $e_i$ would be deleted from the
array in the digital standard $n$-simplex $\Delta^n$.

\begin{Definition} {\rm For each $n$ and $i$, we now define the {\it $i$th face function}
$$
\epsilon_i = \epsilon_i^n : \Delta^{n-1} \longrightarrow \Delta^n
$$
as the function which would send the ordered vertices $\{e_0, \ldots, e_{n-1} \}$ to the ordered vertices $\{e_0, \ldots, \hat {e_i}, \ldots, e_{n} \} \subsetneq \{e_0, \ldots, e_i, \ldots, e_{n} \}$ preserving the displayed orderings as follows:
\begin{itemize}
\item $\epsilon_0^n : (r_0, r_1, \ldots, r_{n-1}) \longmapsto (0, r_0, r_1, \ldots, r_{n-1})$; and
\item $\epsilon_i^n : (r_0, r_1, \ldots, r_{n-1}) \longmapsto (r_0, \ldots,  r_{i-1}, 0, r_i, \ldots, r_{n-1})$ for $i \geq 1$.
\end{itemize}
}\end{Definition}

For example, there are three face functions $\epsilon_i^2 : \Delta^1
\rightarrow \Delta^2$ such as $\epsilon_0^2 : \{ e_0, e_1\}
\rightarrow  \{ e_1, e_2\}$; $\epsilon_1^2 : \{ e_0, e_1\}
\rightarrow  \{ e_0, e_2\}$; and $\epsilon_2^2 : \{ e_0, e_1\}
\rightarrow  \{ e_0, e_1\}$.

\begin{Definition} {\rm  Let $(X,k_X)$ be a digital image with $k_X$-adjacent relation, and let $\sigma : (\Delta^n, k_{\Delta^n})$ $\rightarrow (X,k_X)$ be a digital $n$-simplex in $(X,k_X)$. Then, the map $\partial_n : dC_{n} (X;R) \rightarrow dC_{n-1} (X;R)$ defined as
$$
\partial_n \sigma =
\begin{cases} {\displaystyle \sum_{i=0}^n} (-1)^i \sigma \circ \epsilon_i^n &\text{if $n \geq 1$}; \\
         0         &\text{if $n =0$. }
\end{cases}
$$
is called the {\it digital boundary operator} of the digital image $(X,k_X)$.
}\end{Definition}

It can be seen that $\partial_n : dC_{n} (X;R) \rightarrow dC_{n-1} (X;R)$ is
an $R$-module homomorphism.
%We thus extend the above definition by linearity to
%the digital singular $n$-chains.
In particular, if $X = \Delta^n$ and $1_{\Delta^n} : \Delta^n \rightarrow \Delta^n$ is the identity, then
$
\partial_n (1_{\Delta^n}) = {\sum_{i=0}^n} (-1)^i \epsilon_i^n.
$
Moreover, if $k < j$, then we obtain
\begin{eqnarray} \label{0}
\epsilon_j^{n+1} \circ \epsilon_k^{n} = \epsilon_k^{n+1} \circ
\epsilon_{j-1}^{n} : \Delta^{n-1} \longrightarrow \Delta^{n+1}.
\end{eqnarray}

\begin{prop} \label{Thm0} For all $n \geq 0$, we have $\partial_n \circ \partial_{n+1} = 0$.
\end{prop}

\begin{proof} It suffices to show that the equation holds for each basis of the free $R$-module, i.e., digital $(n+1)$-simplex
$\sigma : (\Delta^{n+1}, k_{\Delta^{n+1}}) \rightarrow (X,k_X)$ because we can extend it through linearity to
the digital $n$-chains. We have
\begin{eqnarray} \label{1}
\begin{array}{ll}
\partial_n \circ \partial_{n+1} (\sigma)
   & = \partial_n ( {\displaystyle \sum_{j=0}^{n+1}} (-1)^j \sigma \circ \epsilon_j^{n+1} ) \\
   & = \partial_n (\sigma \circ \epsilon_0^{n+1} - \sigma \circ \epsilon_1^{n+1} + \sigma \circ \epsilon_2^{n+1} + \cdots + (-1)^{n+1} \sigma \circ \epsilon_{n+1}^{n+1} ) \\
%   & = {\displaystyle \sum_{k=0}^{n}} (-1)^k \sigma \circ \epsilon_{0}^{n+1}\circ \epsilon_{k}^{n}
%       -{\displaystyle \sum_{k=0}^{n}} (-1)^k \sigma \circ \epsilon_{1}^{n+1}\circ \epsilon_{k}^{n} +\\
%   & ~~~~~~~~~~~~~~~~~~~~~~~~~~~~~~~
% \cdots + (-1)^{n+1} {\displaystyle \sum_{k=0}^{n}} (-1)^k \sigma \circ \epsilon_{n+1}^{n+1}\circ \epsilon_{k}^{n} \\
   & = {\displaystyle \sum_{j=0}^{n+1}} (-1)^j (  {\displaystyle \sum_{k=0}^{n}} (-1)^k
        \sigma \circ \epsilon_j^{n+1} \circ \epsilon_k^{n}) \\
   & = {\displaystyle \sum_{j=0}^{n+1}} {\displaystyle \sum_{k=0}^{n}} (-1)^{j+k}
        \sigma \circ \epsilon_j^{n+1} \circ \epsilon_k^{n} \\
   & = {\displaystyle \sum_{j \leq k}} (-1)^{j+k} \sigma \circ \epsilon_j^{n+1} \circ \epsilon_k^{n}
       + {\displaystyle \sum_{k < j}} (-1)^{j+k} \sigma \circ \epsilon_j^{n+1} \circ \epsilon_k^{n} \\
   & = {\displaystyle \sum_{j \leq k}} (-1)^{j+k} \sigma \circ \epsilon_j^{n+1} \circ \epsilon_k^{n}
       + {\displaystyle \sum_{k < j}} (-1)^{j+k} \sigma \circ \epsilon_k^{n+1} \circ
       \epsilon_{j-1}^{n} .
\end{array}
\end{eqnarray}
The left-hand and the right-hand terms in the last two summations in (\ref{1}) can be
expressed by the upper triangular region and the lower triangular
region, respectively, in the $(n+2) \times (n+1)$-matrix-like form
as follows:
\renewcommand{\tabcolsep}{16.48pt}
\renewcommand{\arraystretch}{1.20}
\begin{table}[h!]
  \begin{center}
    \caption{When $j \leq k$}
    \label{tab:table1}
\begin{tabular}{|c|cccc|}
    \hline
\cellcolor{lightgray}$j \leq k$ & \cellcolor{lightgray}$k=0$   &  \cellcolor{lightgray} $k=1$  & \cellcolor{lightgray} $\ldots$  & \cellcolor{lightgray} $k=n$  \\
    \hline
$j=0$   &$~~~ \sigma \epsilon_0^{n+1}\epsilon_0^{n}~~~$  &$~~~-
\sigma \epsilon_0^{n+1}\epsilon_1^{n}$~~~
        &~~~$\ldots$~~~ &~~~$(-1)^n \sigma \epsilon_0^{n+1} \epsilon_n^{n}$~~~ \\
$j=1$   &$0$     &$\sigma \epsilon_1^{n+1}\epsilon_1^{n}$
        &$\ldots$  &$(-1)^{n+1}\sigma \epsilon_1^{n+1}\epsilon_n^{n}$   \\
$j=2$   &$0$    &$0$
        &$\ldots$  &$(-1)^{n+2}\sigma \epsilon_2^{n+2}\epsilon_n^{n}$   \\
$\vdots$ & $\vdots$    & $\vdots$  & $\ddots$ & $\vdots$  \\
$j=k$   &$0$     &$0$  &$\ldots$  &$(-1)^{n+j}\sigma \epsilon_j^{n+1}\epsilon_n^{n}$   \\
$\vdots$ & $\vdots$    & $\vdots$  & $\ddots$ & $\vdots$  \\
$j=n$    &$0$  &$0$
        &$\ldots$  &$(-1)^{2n}\sigma \epsilon_n^{n+1}\epsilon_n^{n}$   \\
$j=n+1$ &$0$ &$0$ &$\ldots$ &$0$   \\
    \hline
\end{tabular}
\end{center}
\end{table}

\renewcommand{\tabcolsep}{14.48pt}
\renewcommand{\arraystretch}{1.20}
\begin{table}[h!]
  \begin{center}
    \caption{When $k < j$}
    \label{tab:table2}
\begin{tabular}{|c|cccc|}
    \hline
\cellcolor{lightgray}$k <j$   &\cellcolor{lightgray}$k=0$   &\cellcolor{lightgray}  $k=1$  &\cellcolor{lightgray} $\ldots$
&\cellcolor{lightgray} $k=n$  \\
    \hline
$j=0$   &$0$ &$0$ &$\ldots$ &$0$  \\
$j=1$   &$-\sigma \epsilon_1^{n+1}\epsilon_0^{n}$ &$0$ &$\ldots$ &$0$  \\
$j=2$   &$\sigma \epsilon_2^{n+1}\epsilon_0^{n}$  &$-\sigma \epsilon_2^{n+1}\epsilon_1^{n}$  &$\ldots$ &$0$   \\
& $\vdots$    & $\vdots$  & $\ddots$ & $\vdots$  \\
$j=k$  &$(-1)^{j}\sigma \epsilon_j^{n+1}\epsilon_0^{n}$
&$(-1)^{j+1}\sigma \epsilon_j^{n+1}\epsilon_1^{n}$
         &$\ldots$  &$0$  \\
& $\vdots$    & $\vdots$ & $\ddots$ & $\vdots$  \\
$j=n$ &$(-1)^{n}\sigma \epsilon_n^{n+1}\epsilon_0^{n}$
&$(-1)^{n+1}\sigma \epsilon_n^{n+1}\epsilon_1^{n}$
        &$\ldots$  &$0$   \\
$j=n+1$ &$(-1)^{n+1}\sigma \epsilon_{n+1}^{n+1}\epsilon_0^{n}$
&$(-1)^{n+2}\sigma \epsilon_{n+1}^{n+1}\epsilon_1^{n}$
        &$\ldots$  &$(-1)^{2n+1}\sigma \epsilon_{n+1}^{n+1}\epsilon_n^{n}$ \\
    \hline
\end{tabular}
\end{center}
\end{table}

%\vspace{0.4cm}

\noindent By using equation (\ref{0}), we can rewrite Table \ref{tab:table2} as
follows:
\renewcommand{\tabcolsep}{13.48pt}
\renewcommand{\arraystretch}{1.20}
\begin{table}[h!]
  \begin{center}
    \caption{When $k < j$}
    \label{tab:table3}
\begin{tabular}{|c|cccc|}
    \hline
\cellcolor{lightgray}$k <j$  &\cellcolor{lightgray}$k=0$   &\cellcolor{lightgray}$k=1$  &\cellcolor{lightgray} $\ldots$
&\cellcolor{lightgray}$k=n$  \\
    \hline
$j=0$   &$0$ &$0$ &$\ldots$ &$0$  \\
$j=1$   &$-\sigma \epsilon_0^{n+1}\epsilon_0^{n}$ &$0$ &$\ldots$ &$0$  \\
$j=2$   &$\sigma \epsilon_0^{n+1}\epsilon_1^{n}$  &$-\sigma \epsilon_1^{n+1}\epsilon_1^{n}$  &$\ldots$ &$0$   \\
& $\vdots$    & $\vdots$  & $\ddots$ & $\vdots$  \\
$j=k$  &$(-1)^{j}\sigma \epsilon_0^{n+1}\epsilon_{j-1}^{n}$
&$(-1)^{j+1}\sigma \epsilon_1^{n+1}\epsilon_{j-1}^{n}$
         &$\ldots$  &$0$  \\
& $\vdots$    & $\vdots$ & $\ddots$ & $\vdots$  \\
$j=n$ &$(-1)^{n}\sigma \epsilon_0^{n+1}\epsilon_{n-1}^{n}$
&$(-1)^{n+1}\sigma \epsilon_1^{n+1}\epsilon_{n-1}^{n}$
        &$\ldots$  &$0$   \\
$j=n+1$ &$(-1)^{n+1}\sigma \epsilon_{0}^{n+1}\epsilon_n^{n}$
&$(-1)^{n+2}\sigma \epsilon_{1}^{n+1}\epsilon_n^{n}$
        &$\ldots$  &$(-1)^{2n+1}\sigma \epsilon_{n}^{n+1}\epsilon_n^{n}$ \\
    \hline
\end{tabular}
\end{center}
\end{table}

%\vspace{0.4cm}

\noindent The terms in Tables \ref{tab:table1} and \ref{tab:table3} cancel in pairs. Thus,
$$
\partial_n \circ \partial_{n+1} (\sigma) = 0
$$
for each digital $(n+1)$-simplex $\sigma : (\Delta^{n+1}, k_{\Delta^{n+1}}) \rightarrow (X,k_X)$,
as required.
\end{proof}

\begin{Definition} {\rm The kernel of $\partial_n : dC_{n} (X;R) \rightarrow dC_{n-1} (X;R)$ is called the {\it module of digital $n$-cycles} in $(X,k_X)$ and denoted by $dZ_n (X;R)$. The image of $\partial_{n+1} : dC_{n+1} (X;R) \rightarrow dC_{n} (X;R)$ is called the {\it module of digital $n$-boundaries} in $(X,k_X)$ and denoted by $dB_n (X;R)$.
}\end{Definition}

By Proposition \ref{Thm0}, each digital $n$-boundary of digital $(n+1)$-chains is automatically a digital
$n$-cycle; that is,  $dB_n (X;R)$ is a submodule of $dZ_n(X;R)$ for each $n \geq 0$. Thus, we can define the following:

\begin{Definition} {\rm For each $n \geq 0$, the {\it $n$th digital homology module} $dH_n  (X;R)$ over $R$ of a digital image $(X,k_X)$ with $k_X$-adjacent relation is defined by
$$
dH_n  (X;R) =  dZ_n (X;R) /  dB_n (X;R).
$$
The coset $[z_n] = z_n +  B_n (X;R)$ is called the {\it digital homology class} of $z_n$, where $z_n$ is a digital $n$-cycle.
The {\it $n$th digital cohomology module} $dH^n  (X;R)$ over $R$ of a digital image $(X,k_X)$ is defined as the corresponding cohomology of the cochain complex over $R$ obtained by the dual modules over $R$ and dual $R$-module homomorphisms.

}\end{Definition}

We note that if $f : (X,k_X) \rightarrow (Y,k_Y)$ is
$(k_X,k_Y)$-continuous and if $\sigma : (\Delta^n, k_{\Delta^n})
\rightarrow (X,k_X)$ is a digital $n$-simplex in $(X,k_X)$,
then
$$
f \circ \sigma : (\Delta^n, k_{\Delta^n}) \rightarrow (Y,k_Y)
$$
is a digital $n$-simplex in $(Y,k_Y)$. By extending through linearity, we
have an $R$-module homomorphism
$$
f_\sharp : dC_n (X;R) \rightarrow  dC_n (Y;R)
$$
defined by
$$
f_\sharp (\Sigma r_\sigma \sigma ) = \Sigma r_\sigma (f \circ
\sigma),
$$
where $r_\sigma$ is an element of the ring $R$.
Moreover, the following diagram
$$
\xymatrix@C=10mm @R=7mm{
 dC_n (X;R) \ar[r]^-{f_\sharp} \ar[d]^{\partial_n} & dC_n (Y;R)  \ar[d]^{\partial_n'} \\
 dC_{n-1} (X;R) \ar[r]^-{f_\sharp}  & dC_{n-1} (Y;R), }
$$
commutes for every $n \geq 0$, where $\partial_n'$ is a digital boundary operator of the digital image $(Y,k_Y)$.

%Let $\mathcal A$ and $\mathcal B$ be categories. We recall that a {\it covariant functor} $S : \mathcal A \rightarrow \mathcal B$ consists of an object %function which assigns to every object $X$ of $\mathcal A$ an object $S(X)$ of $\mathcal B$ and a morphism function which assigns to every morphism $f %: X \rightarrow Y$ of $\mathcal A$ a morphism $S(f) : S(X) \rightarrow S(Y)$ of $\mathcal B$
%such that $S(1_X ) = 1_{S(X)}$, and $S(g\circ f) = S(g) \circ S(f)$.

Let $\mathcal D$ be the category of digital images and digital
continuous functions, and $\mathcal M$ be the category of
$R$-modules and $R$-module homomorphisms, then, we have the following
(see \cite[Theorem 3.15] {BKO} for the case of digital simplicial
homology groups).

\begin{prop} \label{Thm-1} For each $n \geq 0$,  $dH_n : \mathcal D \rightarrow \mathcal M$ is a covariant functor, and $dH^n : \mathcal D \rightarrow \mathcal M$ is a contravariant functor.
\end{prop}

\begin{proof} If  $f : (X,k_X) \rightarrow (Y,k_Y)$ is a $(k_X,k_Y)$-continuous function, then we define
$$
f_* =   dH_n (f) :   dH_n (X;R) \rightarrow   dH_n (Y;R)
$$
by
$$
dH_n (f) ([z_n]) = f_\sharp (z_n ) +   dB_n (Y;R),
$$
where $z_n \in dZ_n (X;R)$. We note that if $z_n \in dZ_n (X;R)$, then $\partial_n (z_n) = 0$ and thus
$$
\partial_n' \circ f_\sharp (z_n) = f_\sharp \circ \partial_n (z_n ) = 0;
$$
that is,
$$
f_\sharp (z_n) \in {\rm ker}~ \partial_n' = dZ_n (Y;R).
$$
Moreover, if $b_n \in dB_n (X;R)$, then
$$
\partial_{n+1} (b_{n+1}) = b_n
$$
for some $b_{n+1} \in  dC_{n+1} (X;R)$, and
$$
f_\sharp (b_n ) = f_\sharp \circ \partial_{n+1} (b_{n+1}) =
\partial_{n+1}' \circ f_\sharp (b_{n+1}) \in   dB_n (Y;R).
$$
If $b_n \in   dB_n (X;R)$, then
$$
f_\sharp (z_n + b_n ) +   dB_n (Y;R) = f_\sharp (z_n) +  dB_n (Y;R);
$$
that is, the definition of $dH_n (f)$ is independent of the choice
of representatives.

If $1_X : (X, k_X) \rightarrow  (X, k_X)$ is the identity, and
$\sigma : (\Delta^n, k_{\Delta^n}) \rightarrow (X, k_X)$ is a digital $n$-simplex, then
$$
{1_X}_\sharp (\sigma) = 1_X \circ \sigma = \sigma;
$$
that is, $dH_n (1_X) = 1_{dH_n (X;R)}$, the identity automorphism on $dH_n (X;R)$.

If  $f : (X,k_X) \rightarrow (Y,k_Y)$ is digitally
$(k_X,k_Y)$-continuous, and $g : (Y,k_Y) \rightarrow (Z,k_Z)$ is a
digitally $(k_Y,k_Z)$-continuous function, then
$$
(g \circ f)_\sharp (\sigma) = (g \circ f) \circ (\sigma) = g \circ
(f \circ \sigma) = g_\sharp (f_\sharp (\sigma)).
$$
Therefore, we have
$$
dH_n (g \circ f) = dH_n (g) \circ dH_n (f).
$$
Similarly, it can be shown that $dH^n : \mathcal D \rightarrow \mathcal M$ is a contravariant functor in a precisely analogous way.
\end{proof}

\begin{rmk}
Let $(X,k_X)$ be a digital image with $k_X$-adjacent relation, and let $X \times [0,m]_{\mathbb{Z}}$ be a digital image with cartesian product
whose adjacent relation is denoted by $k_C$. Then the map  $\psi_i : X \rightarrow X \times [0,m]_{\mathbb{Z}}$ defined by
$\psi_i (x) = (x,i)$ for $i = 0, m$ is $(k_X, k_C)$-continuous. Moreover,
if
$$
dH_n (\psi_0 ) = dH_n (\psi_m ) : dH_n (X;R) \rightarrow dH_n (X \times [0,m];R)
$$
as $R$-module homomorphisms, and if
$$
f\simeq_{(k_X,k_Y)} g : (X,k_X ) \rightarrow (Y,k_Y),
$$
then we have
$$
dH_n (f) =   dH_n (g ) :  dH_n (X,R) \rightarrow  dH_n(Y;R).
$$
\end{rmk}

Let $i_1: X \rightarrow X \times X$ be the first inclusion, and $i_2 : X \rightarrow X \times X$ be the second inclusion, then, from Proposition \ref{Thm-1}, we have $R$-module homomorphisms
$$
{i_1}_*, {i_2}_* : dH_* (X; R) \rightarrow dH_* (X \times X; R)
$$
of digital homology modules induced by $i_1$ and $i_2$, respectively.

\begin{Definition} {\rm
An element $x \in dH_* (X; R)$ is said to be a {\it digital primitive homology class} if
$\Delta_* (x) = {i_1}_*(x) + {i_2}_*(x)$, where $\Delta : X \rightarrow X \times X$ is the diagonal map.
}\end{Definition}

Let $PdH_* (X; R)$ denote the submodule of $dH_* (X; R)$ with coefficients in a commutative ring $R$ consisting of all the digital primitive homology classes. Then, we have the following.

\begin{prop} \label{prop1}
Let $f : X \rightarrow Y$ be a $(k_X, k_Y)$-continuous function. Then
$$
f_* (PdH_* (X; R)) \subseteq PdH_* (Y; R).
$$
\end{prop}

\begin{proof}
If $x$ is any digital primitive homology class of $dH_* (X; R)$ with coefficients in a commutative ring $R$, then from the commutative diagram
$$
\xymatrix@C=10mm @R=8mm{
dH_* (X; R) \ar[r]^-{f_*} \ar[d]_-{\Delta_*}  &dH_* (Y; R)  \ar[d]^-{\Delta_*} \\
dH_* (X \times X; R) \ar[r]^-{(f \times f)_*} &dH_* (Y \times Y; R),
}
$$
we have
$$
\begin{array}{lll}
\Delta_* (f_* (x)) &= (f \times f)_* \circ \Delta_* (x) \\
&= (f \times f)_* ({i_1}_* (x) + {i_2}_* (x)) \\
&= {i_1}_* (f_* (x)) + {i_2}_* (f_* (x)), \\
\end{array}
$$
where $i_1 : W \rightarrow W \times W$ is the first inclusion and $i_2 : W \rightarrow W \times W$ is the second inclusion on $W = X$ or $Y$;
that is, the homomorphic image of the digital primitive homology classes is also digital primitive, as required.
\end{proof}

The following result is the digital counterpart of the dimension axiom as one of the Eilenberg-Steenrod axioms in classical algebraic topology.

\begin{prop} If $(X,k_X)$ is a one-point digital image with $k_X$-adjacent relation, then
$$
dH_n (X;R) =
\begin{cases}  0               &\text{ for all $n \geq 1$}; \\
         R                &\text{ $n =0$. }
\end{cases}
$$
\end{prop}

\begin{proof} For each $n \geq 0$, we have
$
dC_n (X;R) ~\cong~ R ~\cong~ <\sigma_n >
$
generated by the digital $n$-simplex $\sigma_n : (\Delta^n, k_{\Delta^n})
\rightarrow (X,k_X)$ which is a constant function in this case. From
the definition of the boundary operator
$\partial_n :  dC_n (X;R) \rightarrow  dC_{n-1} (X;R)$, we obtain
$$
\partial_n ( \sigma_n )=
\begin{cases} 0,               &\text{ if $n$ is odd}; \\
       \sigma_{n-1},           &\text{ if $n$ is even}:
\end{cases}
$$
Thus, $\partial_n$ is a trivial homomorphism if $n$ is odd, and if $n$ is even, it
is an isomorphism.

Assuming $n \geq 1$, and considering the chain complex of $R$-modules
$$
\xymatrix{ \cdots \ar[r]  & dC_{n+1} (X;R)  \ar[r]^-{\partial_{n+1}}
& dC_{n} (X;R) \ar[r]^-{\partial_{n}} & dC_{n-1} (X;R) \ar[r] &\cdots. }
$$
\begin{enumerate}
\item[\rm{(1)}] If $n \geq 1$ is odd, then  $\partial_n$ is a trivial homomorphism and $\partial_{n+1}$ is an isomorphism, and thus,
$$
dC_n (X;R) = {\rm ker}~ \partial_n = dZ_n (X;R)
$$
and
$$
dC_n (X;R) ={\rm im}~ \partial_{n+1} = dB_n (X;R).
$$
Therefore, we have
$$
dH_n (X;R) = dZ_n (X;R) /dB_n (X;R) = dC_n (X;R) /dC_n (X;R) = 0.
$$
\item[\rm{(2)}] Similarly, if $n \geq 2$ is even, then  $\partial_n$ is an isomorphism, and thus
$$
Z_n (X,R) = {\rm ker}~ \partial_n = 0,
$$
so that
$$
dH_n (X;R) =  dZ_n (X;R) /dB_n (X;R) = 0 /dB_n (X;R) = 0.
$$
\end{enumerate}
On the other hand, from the sequence
$$
\xymatrix{ \cdots \ar[r]  & dC_{1} (X;R)  \ar[r]^-{\partial_{1}} &
dC_{0} (X;R) \ar[r]^-{\partial_{0}} &0  , }
$$
we have
$$
\begin{cases}
{\rm ker}~ \partial_0 = dZ_0 (X;R) \cong R ; ~~{\rm and} \\
{\rm im}~ \partial_1 =  dB_0 (X;R) \cong 0,
\end{cases}
$$
so that
$$
dH_0 (X;R) =  dZ_0 (X;R) /dB_0 (X;R) \cong R,
$$
as required.
\end{proof}

\begin{Definition} {\rm
Given a digital image $(X, k_X)$, we define an equivalence relation on
$(X,k_X)$ by setting $ x \sim y$, if there is a $k_X$-connected digital
subimage containing both $x$ and $y$. The equivalence classes are
called the {\it $k_X$-connected components} of $(X,k_X)$.
}\end{Definition}

\begin{prop} Let $\{(X_\lambda ,k_X) | \lambda \in \Lambda \}$ be the set of $k_X$-connected components of a digital image $(X,k_X)$. Then,
$$
dH_n (X;R) \cong \bigoplus_{\lambda \in \Gamma} dH_n (X_\lambda;R)
$$
for every $n \geq 0$.
\end{prop}

\begin{proof} If $d = \sum r_i \sigma_i \in dC_n (X;R)$,
then the digital continuity of the digital $n$-simplex shows that
each ${\rm im} (\sigma_i)$ is contained in a unique $k_X$-connected
component of $(X,k_X)$. Thus we may write
$
c = \sum_{\lambda \in \Lambda} c_\lambda ,
$
where $c_\lambda$ is the sum of those terms; that is, $c_\lambda = \sum r_j \sigma_j^\lambda$ in $c$ for which  ${\rm im} (\sigma_j^\lambda)
\subset X_\lambda$ for each $\lambda \in \Lambda$. Now we define an
isomorphism
$$
\Psi : dC_n (X;R) \rightarrow \bigoplus_{\lambda \in \Lambda} dC_n(X_\lambda;R)
$$
by $c \mapsto (c_\lambda )_{\lambda \in \Lambda}$.
Now $c$ is a cycle if and only if each $c_\lambda$ is a cycle. Since
${\rm im} (\sigma_j^\lambda) \subset X_\lambda$ implies
${\rm im} (\sigma_j^\lambda \circ \epsilon_i ) \subset X_\lambda$, $\partial
\sigma_j^\lambda \in dC_{n-1} (X_\lambda;R)$, and the assumption $0 =
\partial c = \sum \partial c_\lambda$ implies $\partial c_\lambda =0$
for all $\lambda \in \Lambda$. Indeed, an element in the direct sum
$\bigoplus_{\lambda \in \Lambda} dC_{n-1} (X_\lambda;R)$ is zero if and
only if its coordinates are all zero. It follows that the map
$$
\psi : dH_n (X;R) \rightarrow \bigoplus_{\lambda \in \Lambda} dH_n(X_\lambda;R)
$$
given by
$[c] \mapsto ([c_\lambda ] )_{\lambda \in \Lambda}$
is well defined.

We now define a map
$$
\varphi : \bigoplus_{\lambda \in \Lambda} dH_n (X_\lambda;R)
\rightarrow dH_n (X;R)
$$
by
$
([c_\lambda ] )_{\lambda \in \Lambda} \mapsto [  \sum_{\lambda \in
\Lambda} c_\lambda  ].
$
Then, it is easy to see that $\psi$ and $\varphi$ are the mutual
inverses as required.
\end{proof}

\bigskip

\section{Digital Hopf spaces, Pontryagin algebras and coalgebras}\label{R}

From many kinds of algebraic structures, we can think of a Hopf group in algebraic topology as generalization of a usual group in algebra (see \cite{S} and \cite{W}). For application in computer science, the notions of Hopf spaces or Hopf groups in algebraic topology will be transformed in this section to those of digital theoretical counterparts (compare with \cite{EK} and \cite{EK1}).

\begin{Definition} {\rm (\cite{L8, L11})
Let $e_{y_0} : Y \rightarrow Y$ be a constant function at $y_0$ and let $1_Y : Y \rightarrow Y$ be an identity function on $Y$.
A {\it digital Hopf space} $Y = (Y, y_0, k_Y, m_Y)$ consists of a pointed digital image $(Y, y_0)$ with an adjacent relation $ k_Y$ and
a $(k_{Y \times Y}, k_Y)$-continuous function
$m_Y : Y \times Y \rightarrow Y$
so that the following diagram
$$
\xymatrix@C=15mm @R=8mm{
Y \ar[dr]_-{1_Y} \ar[r]^-{(e_{y_0}, 1_Y)} & Y \times Y \ar[d]_-{m_Y} &Y \ar[l]_-{(1_Y, e_{y_0})} \ar[dl]^-{1_Y}  \\
&Y }
$$
is commutative up to pointed digital homotopy.
Here, $k_{Y \times Y}$ is an adjacent relation on $Y\times Y$, and
$(e_{y_0}, 1_Y) : Y \rightarrow Y \times Y$
is the composite of the diagonal function $\Delta : Y \rightarrow Y \times Y$ with the product of functions
$(e_{y_0} \times 1_Y) : Y \times Y \rightarrow Y \times Y$,
and similarly for $(1_Y, e_{y_0})$. In this case,
the $(k_{Y \times Y}, k_Y)$-continuous function $m_Y : Y \times Y \rightarrow Y$ above is called a {\it digital multiplication} on $(Y, y_0, k_Y, m_Y)$, and $e_{y_0}$ is called a {\it digital homotopy identity}.
}\end{Definition}

As usual, we denote the pointed digital homotopy class by $[f]$ as the equivalence class of a pointed digital continuous function
$f : (X, x_0, k_X) \rightarrow (Y, y_0, k_Y)$.

\begin{Definition}\label{Def} {\rm  (\cite{L8, L11})
Let $(Y,y_0, k_Y, m_Y)$ be a digital Hopf space with a digital multiplication $m_Y : Y \times Y \rightarrow Y$.
For pointed $(k_X,k_Y)$-continuous functions $f, g : (X,x_0) \rightarrow (Y,y_0)$, we define a {\it digital convolution} $[f]\boxplus[g]$ between pointed digital homotopy classes $[f]$ and $[g]$ by the pointed digital homotopy class of the following compositions
$$
\xymatrix@C=10mm @R=10mm{
f \boxplus g : X \ar[r]^-{\Delta} & X \times X  \ar[r]^-{f \times g} & Y \times  Y \ar[r]^-{m_Y} & Y; }
$$
that is,
$
[f]\boxplus[g] = [m_Y (f \times g) \Delta] = [f\boxplus g],
$
where $\Delta$ is a diagonal function.
}\end{Definition}

\begin{thm}\label{Thm1} Let $(Y,y_0, k_Y, m_Y)$ be a digital Hopf space with a digital multiplication $m_Y : Y \times Y \rightarrow Y$.
If $f,g : (X,x_0) \rightarrow (Y,y_0)$ are $(k_X, k_Y)$-continuous functions and $x \in PdH(X;R)$, then
$$
([f]\boxplus[g])_* (x) = f_* (x) + g_* (x),
$$
where $f_*, g_* : dH_* (X; R) \rightarrow dH_* (Y; R)$ are homomorphisms of digital homology modules over $R$ induced by $(k_X,k_Y)$-continuous functions $f$ and $g$, respectively.
\end{thm}

\begin{proof} We note that $f \boxplus g \simeq_{(k_X,k_Y)} m_Y \circ (f \times g) \circ \Delta$ and the following diagram
$$
\xymatrix@C=10mm @R=8mm{
dH_*(X;R) \ar[r]^-{{i_1^X}_*} \ar[d]_-{f_*}  &dH_*(X\times X;R)  \ar[d]^-{(f \times g)_*} \\
dH_*(Y;R) \ar[r]^-{{i_1^Y}_*} &dH_*(Y \times Y;R),
}
$$
is strictly commutative and similarly for ${i_2^X}_* : dH_*(X;R) \rightarrow dH_*(X \times X;R)$ together with
${i_2^Y}_* : dH_*(Y;R) \rightarrow dH_*(Y \times Y;R)$  in digital homology $R$-modules. Since $Y$ has the digital Hopf structure and $x$ is a digital primitive homology class, we have
$$
\begin{array}{lll}
(f \boxplus g)_* (x)   &= (m_Y \circ (f \times g) \circ \Delta)_* (x)\\
                       &= {m_Y}_* \circ (f \times g)_* \circ \Delta_* (x)\\
                       &= {m_Y}_* \circ (f \times g)_* ({i_1^X}_*(x) + {i_2^X}_*(x))\\
                       &= {m_Y}_*  ({i_1^Y}_* \circ f_*(x) + {i_2^Y}_* \circ g_* (x))\\
                       &= ({m_Y}_*  \circ {i_1^Y}_*) f_*(x) + ({m_Y}_* \circ {i_2^Y}_*) g_* (x)\\
                       &= {1_Y}_* \circ f_*(x) + {1_Y}_* \circ g_* (x)  \\
                       &= f_*(x) +  g_* (x),\\
\end{array}
$$
where $1_Y : Y \rightarrow Y$ is the identity map.
\end{proof}

The following shows that the digital Hopf spaces are closed under the digital products of digital images.

\begin{thm} \label{Thm11}
If $(X,x_0, k_X, m_X)$ and $(Y,y_0, k_Y, m_Y)$ are digital Hopf spaces, then $(X\times Y, x_0 \times y_0, k_{X\times Y}, m_{X\times Y})$ is a digital Hopf space.
\end{thm}

\begin{proof}
Let $m_X : X \times X \rightarrow X$ and $m_Y : Y \times Y \rightarrow Y$ be the digital multiplications on $X$ and $Y$, respectively. We define a function
$$
m_{X\times Y} : (X \times Y) \times  (X \times Y) \longrightarrow  X \times Y
$$
by making the following diagram commute:
$$
\xymatrix@C=15mm @R=10mm{
(X \times Y) \times  (X \times Y) \ar[d]_-{1_X \times S_{Y\times X} \times 1_Y} \ar[r]^-{m_{X\times Y}} & X \times Y \\
(X \times X) \times  (Y \times Y) \ar[ur]_-{m_X \times m_Y}, }
$$
where $S_{Y\times X} : Y \times X \rightarrow X \times Y$ is a switching function, and $1_X$ and $1_Y$ are the identity functions on $X$ and $Y$, respectively. Then, for all $(x,y) \in X \times Y$, we have
$$
\begin{array}{lll}
m_{X\times Y} (1_{X \times Y}, c_{x_0 \times y_0}) (x,y)
   &= m_{X\times Y} (1_{X \times Y} \times c_{x_0 \times y_0}) \Delta (x,y) \\
   &= m_{X\times Y}(x,y,x_0,y_0) \\
   &= (m_X \times m_Y)(1_X \times S_{Y\times X} \times 1_Y)(x,y,x_0,y_0) \\
   &= (m_X \times m_Y)(x,x_0, y,y_0) \\
   &= (m_X (x,x_0), m_Y (y,y_0)) \\
   &= (m_X (1_X \times c_{x_0}) \Delta (x), m_Y (1_Y \times c_{y_0}) \Delta (y)) \\
   &= (m_X (1_X, c_{x_0}) (x), m_Y (1_Y, c_{y_0}) (y)) \\
   &\simeq_{(k_{X \times Y},k_{X \times Y})} (1_X (x), 1_Y (y)) = (x,y) \\
   &= 1_{X\times Y} (x,y);\\
\end{array}
$$
that is,
$$
m_{X\times Y} (1_{X \times Y}, c_{x_0 \times y_0}) \simeq_{(k_{X \times Y},k_{X \times Y})} 1_{X\times Y}.
$$
Similarly, we also obtain
$$
m_{X\times Y} (c_{x_0 \times y_0}, 1_{X \times Y}) \simeq_{(k_{X \times Y},k_{X \times Y})} 1_{X\times Y}.
$$
Therefore,
$$
m_{X\times Y} : (X \times Y) \times  (X \times Y) \rightarrow X \times Y
$$
is a digital multiplication, that is, $(X\times Y, x_0 \times y_0, k_{X\times Y}, m_{X\times Y})$ is a digital Hopf space.
\end{proof}

\begin{cor} Let $(Y,y_0, k_Y, m_Y)$ be a digital Hopf space and let $i_1, i_2 : Y \rightarrow Y \times Y$ be the first and second inclusions, respectively. Then we obtain
$$
(i_1 \boxplus i_2)_* = \Delta_*,
$$
and, in particular,
$$
(i_1 \boxplus i_2)_* (PdH_* (Y;R)) = \Delta_* (PdH_* (Y;R)).
$$
%the image of the $R$-module homomorphism $\Delta_* : dH_* (Y;R) \rightarrow dH_* (Y \times Y;R)$ induced by the diagonal map $\Delta : Y \rightarrow Y \times %Y$  is the same as the primitive subspace $PdH_* (Y;R)$ of $dH_* (Y;R)$.
\end{cor}

\begin{proof}
Let $m_Y : Y \times Y \rightarrow Y$ be the digital multiplication on $(Y,y_0)$. Then, from Theorem \ref{Thm11}, $Y \times Y$ is also a digital Hopf space with a digital multiplication $m_{Y \times Y}$. Since $m_Y \circ i_1 \simeq_{(k_Y,k_Y)} 1_Y$ and $m_Y \circ i_2 \simeq_{(k_Y,k_Y)} 1_Y$, we have
$$
\begin{array}{lll}
[i_1 \boxplus i_2]
   &= [m_{Y\times Y} (i_1 \times i_2) \Delta] \\
   &= [(m_Y \times m_Y)\circ(1_Y \times S_{Y\times Y} \times 1_Y)\circ (i_1 \times i_2) \Delta] \\
   &= [(m_Y \times m_Y)\circ (i_1 \times i_2) \Delta] \\
   &= [((m_Y \circ i_1) \times (m_Y \times i_2)) \Delta] \\
   &= [(1_Y \times 1_Y)\Delta] \\
   &= [\Delta], \\
\end{array}
$$
where $S_{Y \times Y} : Y \times Y \rightarrow Y \times Y$ is the switching function.
If $y$ is any digital primitive homology class, then, from Theorem \ref{Thm1}, we obtain
$$
(i_1 \boxplus i_2)_* (y)  = \Delta_* (y) =  {i_1}_* (y) + {i_2}_* (y),
$$
as required.
\end{proof}

\begin{Definition} {\rm
Let $(Y, y_0, k_Y, m_Y)$ be a digital Hopf space with a digital multiplication $m_Y : Y \times Y \rightarrow Y$. Then,
an element $y \in dH^* (Y; R)$ is said to be a {\it digital primitive cohomology class} if
$$
{m_Y^*} (y) = {p_1^*} (y) + {p_2^*}(y),
$$
where $p_1^*, p_2^* : dH^*(Y;R) \rightarrow dH^*(Y\times Y;R)$ are the homomorphisms of cohomology $R$-modules induced by
$p_1 , p_2 : Y \times Y \rightarrow Y$, the first and second projections onto $Y$, respectively.
}\end{Definition}

\begin{Definition} {\rm
Let $(X, x_0, k_X, m_X)$ and $(Y, y_0, k_Y, m_Y)$ be pointed digital H-spaces. A $(k_X, k_Y)$-continuous function $f : X \rightarrow Y$ is said to be a {\it digital Hopf function} if the following diagram
$$
\xymatrix@C=15mm @R=8mm{
X \times X \ar[d]_-{m_X} \ar[r]^-{f \times f} & Y \times Y \ar[d]^-{m_Y} \\
X \ar[r]^-{f} &Y }
$$
commutes up to the digital homotopy.
}\end{Definition}

Let $PdH^* (X; R)$ denote the submodule of $dH^* (X; R)$ consisting of all the digital primitive cohomology classes. Then, we have the dual of Proposition \ref{prop1} as follows.

\begin{prop} Let $(X,x_0, k_X, m_X)$ and $(Y,y_0, k_Y, m_Y)$ be digital Hopf spaces, and let $f : X \rightarrow Y$ be a digital Hopf function between digital Hopf spaces. Then
$$
f^* (PdH^* (Y; R)) \subseteq PdH^* (X; R).
$$
\end{prop}

\begin{proof}
Since $f : X \rightarrow Y$ be a digital Hopf function, we have
$$
f \circ m_X \simeq_{(k_{X \times X}, k_Y)}  m_Y \circ (f\times f).
$$
If $y$ is any digital primitive cohomology class of $dH^* (Y; R)$ with coefficients in a commutative ring $R$, then from the commutative diagram
$$
\xymatrix@C=10mm @R=8mm{
dH^* (Y; R) \ar[d]_-{{m_Y}^*} \ar[r]^-{f^*} &dH^* (X; R) \ar[d]^-{{m_X}^*} \\
dH^* (Y \times Y; R) \ar[r]^-{(f \times f)^*}   &dH^* (X\times X; R),   \\
}
$$
we have
$$
\begin{array}{lll}
m_X^* (f^* (y)) &= (f \times f)^* \circ m_Y^* (y) \\
&= (f \times f)^* ({p_1}^* (y) + {p_2}^* (y)) \\
&= {p_1}^* (f^* (y)) + {p_2}^* (f^* (y)), \\
\end{array}
$$
where $p_1 , p_2 : W \rightarrow W \times W$ are the first and second projections, respectively, onto $W$ ($= X$ or $Y$);
that is, the homomorphic image of the digital primitive cohomology classes is also digital primitive, as required.
\end{proof}

The following is a dual of Theorem \ref{Thm1}.

\begin{thm}\label{Thm2} Let $(Y,y_0, k_Y, m_Y)$ be a digital Hopf space with a digital multiplication $m_Y : Y \times Y \rightarrow Y$, and let
$f,g : (X,x_0) \rightarrow (Y,y_0)$ be $(k_X, k_Y)$-continuous functions. If $y \in dH(Y;R)$ is digital primitive cohomology class, then
$$
([f]\boxplus[g])^* (y) = f^* (y) + g^* (y),
$$
where $f^*, g^* : dH^* (Y; R) \rightarrow dH^* (X; R)$ are homomorphisms of digital cohomology $R$-modules induced by the $(k_X,k_Y)$-continuous functions $f$ and $g$, respectively.
\end{thm}

\begin{proof}
Let ${p_1}_W, {p_2}_W : W \times W \rightarrow W$ be the first and second projections, respectively, onto $W$($=X$ or $Y$).
Since $f \boxplus g \simeq_{(k_X,k_Y)} m_Y \circ (f \times g) \circ \Delta$ and the following diagram
$$
\xymatrix@C=10mm @R=8mm{
dH^*(Y;R) \ar[r]^-{{p_1^*}_Y} \ar[d]_-{f^*}  &dH^*(Y\times Y;R)  \ar[d]^-{(f \times g)^*} \\
dH^*(X;R) \ar[r]^-{{p_1^*}_X} &dH^*(X \times X;R),
}
$$
is strictly commutative. Similarly, for ${p_2^*}_X : dH^*(X;R) \rightarrow dH^*(X \times X;R)$ together with
${p_2^*}_Y : dH^*(Y;R) \rightarrow dH^*(Y \times Y;R)$  in digital cohomology modules over $R$, we have
$$
\begin{array}{ll}
(f \boxplus g)^* (y)   &= (m_Y \circ (f \times g) \circ \Delta)^* (y)\\
                       &=  \Delta^* \circ (f \times g)^* \circ m_Y^* (y) \\
                       &=  \Delta^* \circ (f \times g)^* ({p_1^*}_Y (y) + {p_2^*}_Y (y))\\
                       &=  \Delta^* \circ ({p_1^*}_X \circ f^* (y) + {p_2^*}_X \circ g^* (y)) \\
                       &=  (\Delta^* \circ {p_1^*}_X ) f^* (y) + (\Delta^* \circ {p_2^*}_X ) g^* (y) \\
                       &= 1_{dH^*(X;R)} \circ f^*(y) + 1_{dH^*(X;R)} \circ g^* (y),\\
                       &= f^*(y) +  g^* (y),\\
\end{array}
$$
where $1_{dH^*(X;R)}: dH^*(X;R) \rightarrow dH^*(X;R)$ is the identity automorphism of $R$-modules, and the third equality is guaranteed by the fact that $y$ is a digital primitive cohomology class.
\end{proof}

\begin{Definition} {\rm
The {\it digital homology cross product} is defined by the homomorphism
$$
\mu_X = \times : dH_* (X; R) \otimes_{R} dH_* (X; R) \rightarrow dH_* (X \times X; R)
$$
of $R$-modules sending $[x] \otimes [y]$ to $[x \times y]$.
}\end{Definition}

Under what conditions can we say that the submodule $PdH_* (X;R) \subseteq dH_* (X;R)$ consisting of digital primitive homology classes is equal to $dH_* (X;R)$? The following gives an answer to this query:

\begin{thm}\label{Thm3}
Let $dH_s (X;R) =0$ for $s \leq n-1$. Then $PdH_s (X;R)$ is equal to $dH_s (X;R)$ for all $s \leq 2n$.
\end{thm}

\begin{proof}
We consider the K\"unneth exact sequence
%$$
%\xymatrix@C=4.5mm @R=0.1mm{
%0 \ar[r]^-{} &\bigoplus_{p+q=m} dH_s(X;R)\otimes dH_t(X;R) \ar[r]^-{\mu} &dH_m (X\times X;R) &\\
%&\ar[r]^-{} &\bigoplus_{p+q=2m-1}{\rm Tor} (dH_s(X;R),dH_{t-1}(X;R)) \ar[r]^-{} &0,&
%}
%$$
\begin{eqnarray} \label{2}
\begin{array}{ll}
0 \longrightarrow  &\bigoplus_{s+t=m} dH_s(X;R)\otimes dH_t(X;R) \longrightarrow dH_m (X\times X;R) \\
&\,\,\,\,\,\,\,\,\,\,\,\,\,\,\,\,\,\,\,\,\,\,\,\,\,\,\,\,\,\, \longrightarrow \bigoplus_{s+t=m}{\rm Tor}^{R} (dH_s(X;R),dH_{t-1}(X;R)) \longrightarrow 0
\end{array}
\end{eqnarray}
of $R$-modules in algebraic topology; see \cite[page 228]{S}. If $m \leq 2n$, then, by our assumption, the torsion part in the above short exact sequence is trivial,
so that the cross product is an isomorphism of $R$-modules. We now consider a homomorphism of $R$-modules
$$
\Delta_* : dH_m(X;R) \rightarrow dH_m(X \times X ;R)
$$
for all $m \geq 0$ induced by the diagonal map
$$
\Delta : X \rightarrow X \times X.
$$
We see that the target of the $R$-module homomorphism $\Delta_*$ is isomorphic to the first term of the K\"unneth exact sequence (\ref{2}). Therefore, for all $x_m \in dH_m(X;R)$, we have
$$
\begin{array}{ll}
\Delta_* (x_m) &= x_m \otimes 1 + x_{m-1} \otimes x_1 +  x_{m-2} \otimes x_2 + \cdots
                 +  x_{2} \otimes x_{m-2}  +  x_{1} \otimes x_{m-1} + 1 \otimes x_m \\
               &= x_m \otimes 1 + x_{m-1} \otimes 0 +  x_{m-2} \otimes 0 + \cdots  +  0 \otimes x_{m-2}  +  0 \otimes x_{m-1} + 1 \otimes x_m \\
               &= x_m \otimes 1 + 1 \otimes x_m \\
               &= {i_1}_* (x_m) + {i_2}_* (x_m) \\
\end{array}
$$
for each $m \leq 2n$; that is, $x_m$ is a digital primitive homology class as required.
\end{proof}

\begin{rmk}
Let $(Y,y_0, k_Y, m_Y)$ be a digital Hopf space with a digital multiplication $m_Y : Y \times Y \rightarrow Y$. Then,
the homomorphism
$$
{m_Y}_* : dH_* (Y \times Y;R) \rightarrow dH_* (Y;R)
$$
between digital homology modules with coefficients in a commutative ring $R$ with identity gives an algebraic structure on the digital homology $dH_*(Y;R)$ which is called the {\it digital Pontryagin algebra} on a digital Hopf space $(Y,y_0)$. Moreover, if the digital homology $R$-module $dH_* (Y;R)$ is free, then the diagonal map $\Delta : Y \rightarrow Y \times Y$ induces a coalgebra structure on the digital homology module $dH_*(Y;R)$.
\end{rmk}

Recall that a {\it graded $R$-algebra} is a graded $R$-module $A = \{ A_m \}$ together with a homomorphism of degree $0$,
$$
m_A : A \otimes A \rightarrow A,
$$
which is called the {\it product} of the graded $R$-algebra. Here, $m_A$ maps $A_m \otimes A_n$ into $A_{m+n}$ for all $m$ and $n$.
The graded $R$-algebra $A$ is said to be {\it associative} if the following diagram
$$
\xymatrix@C=15mm @R=8mm{
 A \otimes A \otimes A \ar[d]_-{m_A \otimes 1_A}  \ar[r]^-{1 \otimes m_A } &  A \otimes A  \ar[d]^-{m_A} \\
A \otimes A  \ar[r]^-{m_A} &Y
}
$$
is commutative.
Let $S : A \otimes A \rightarrow A \otimes A$ be a switching map.
Then the graded $R$-algebra $A$ is said to be {\it commutative} if the following diagram
$$
\xymatrix@C=15mm @R=8mm{
A \otimes A  \ar[dr]_-{m_A}  \ar[rr]^-{S}& &A \otimes A  \ar[dl]^-{m_A} \\
&A
}
$$
commutes with sign in the sense that
$$
m_A(a \otimes b) = (-1)^{s \cdot t} m_A(b \otimes a),
$$
where $a$ and $b$ are homogeneous elements of degree $s$ and $t$, respectively.

\begin{Definition} {\rm (\cite{L8, L11})
A digital multiplication $m_Y : Y \times Y \rightarrow Y$ on a digital Hopf space $(Y, y_0, m_Y, k_Y)$ is said to be {\it digital homotopy associative} if the following diagram
$$
\xymatrix@C=15mm @R=10mm{
Y \times Y \times Y \ar[d]_-{1_Y \times m_Y} \ar[r]^-{m_Y \times 1_Y} & Y \times Y \ar[d]^-{m_Y} \\
Y \times Y \ar[r]^-{m_Y } &Y}
$$
is digital homotopy commutative. A digital Hopf space $(Y, y_0, m_Y, k_Y)$ with a digital multiplication $m_Y : Y \times Y \rightarrow Y$ is said to be a {\it digital homotopy associative Hopf space} if $m_Y$ is digital homotopy associative.
}\end{Definition}

\begin{Definition} {\rm (\cite{L8, L11})
Let $S_{Y\times Y} : Y \times Y \rightarrow Y \times Y$ be a switching function.
A digital multiplication $m_Y : Y \times Y \rightarrow Y$ is said to be {\it digital homotopy commutative} if the following diagram
$$
\xymatrix@C=15mm @R=3mm{
Y \times Y \ar[rr]^-{S_{Y\times Y}} \ar[ddr]_-{m_Y}& &Y \times Y  \ar[ddl]^-{m_Y}\\
&& \\
&Y& }
$$
is commutative up to digital homotopy. A pointed digital Hopf space $(Y, y_0, m_Y, k_Y)$ with a digital multiplication $m_Y : Y \times Y \rightarrow Y$ is said to be a {\it digital homotopy commutative Hopf space} if $m_Y$ is digital homotopy commutative.
}\end{Definition}

For a specific example, see \cite[Example 3.8]{L8}).

We note that if $(Y,y_0, k_Y, m_Y)$ is a digital Hopf space with a digital multiplication $m_Y : Y \times Y \rightarrow Y$, then, by Theorem \ref{Thm-1}, $m_Y$ induces a homomorphism
$$
{m_Y}_* : dH_* (Y \times Y; R) \rightarrow dH_* (Y; R)
$$
of digital homology modules over a commutative ring $R$ with identity $1_R$.

\begin{thm}\label{Thm3}
Let $(Y, y_0, m_Y, k_Y)$ be a digital homotopy associative and commutative Hopf space with a digital multiplication $m_Y : Y \times Y \rightarrow Y$. Then the digital Pontryagin algebra $dH_*(Y;R)$ becomes a non-negatively graded associative and commutative $R$-algebra.
\end{thm}

\begin{proof}
Let
$$
\mu_Y = \times : dH_s(Y;R) \otimes dH_s(Y;R) \rightarrow dH_s(Y \times Y ;R)
$$
be the digital homology cross product. Then, we can consider the composite
\begin{eqnarray} \label{3}
\xymatrix@C=8mm @R=3mm{
dH_s(Y;R) \otimes dH_s(Y;R) \ar[r]^-{\mu_Y} &dH_s(Y \times Y ;R) \ar[r]^-{{m_Y}_*} &dH_s(Y;R)
}
\end{eqnarray}
of $R$-module homomorphisms, which we denote by `$\cdot_P$' or `$\hat {m_Y}_*$'.
Since the digital multiplication $m_Y : Y \times Y \rightarrow Y$ is digital homotopy associative, we can see that the following diagram
$$
\xymatrix@C=23mm @R=8mm{
dH_s(Y;R) \otimes dH_s(Y;R) \otimes dH_s(Y;R) \ar[d]_-{1_{dH_s(Y;R)} \otimes \hat {m_Y}_*} \ar[r]^-{\hat {m_Y}_* \otimes 1_{dH_s(Y;R)}} & dH_s(Y;R) \otimes dH_s(Y;R) \ar[d]^-{\hat {m_Y}_*} \\
dH_s(Y;R) \otimes dH_s(Y;R) \ar[r]^-{\hat {m_Y}_*} &dH_s(Y;R)
}
$$
commutes on digital homology $R$-modules. Moreover, the digital homotopy commutative multiplication $m_Y : Y \times Y \rightarrow Y$ induces the following commutative diagram
$$
\xymatrix@C=8mm @R=2mm{
dH_s(Y;R) \otimes dH_s(Y;R) \ar[rr]^-{S_{*}} \ar[ddr]_-{\hat {m_Y}_*}& & dH_s(Y;R) \otimes dH_s(Y;R)  \ar[ddl]^-{\hat {m_Y}_*} \\
&& \\
&dH_s(Y;R)&
}
$$
on digital homology $R$-modules, where $S_*$ is an $R$-module homomorphism induced by the switching map $S : Y \times Y \rightarrow Y \times Y$. Therefore, $dH_*(Y;R)$ is a graded associative and commutative $R$-algebra.
\end{proof}

Recall that a {\it coalgebra} $C$ over a ring $R$ is an $R$-module together with maps $\varphi: C \rightarrow C \otimes C$ and $ \epsilon : C \rightarrow R$, so that the following diagrams
$$
\xymatrix@C=23mm @R=8mm{
C \ar[d]_-{\varphi} \ar[r]^-{\varphi} & C \otimes C \ar[d]^-{1 \otimes \varphi} \\
C \otimes C \ar[r]^-{\varphi \otimes 1_C} &C \otimes C \otimes C,
}
$$
and
$$
\xymatrix@C=15mm @R=8mm{
C \ar[d]_-{\varphi} \ar[dr]^-{1_C} \ar[r]^-{\varphi} & C \otimes C \ar[d]^-{1_C \otimes \epsilon} \\
C \otimes C \ar[r]^-{\epsilon \otimes 1_C} &R \otimes C \cong C \cong  C \otimes R,
}
$$
are commutative, where $1_C :  C \rightarrow C$ is the identity automorphism. The module homomorphism $\varphi : C \rightarrow C \otimes C$ above is said to be an {\it $R$-algebra comultiplication} on $C$.
It can be shown that $(Y,y_0,k_Y, m_Y)$ is a digital Hopf space with digital multiplication $m_Y : Y \times Y \rightarrow Y$. Then, the digital multiplication $m_Y$ provides the digital cohomology module $dH^*(Y;R)$ with the structure of coalgebra over the commutative ring $R$ with identity $1_R$.

\begin{thm}\label{Thm4}
Let $f : (X,x_0) \rightarrow (Y,y_0)$ be a digital Hopf function between digital Hopf spaces. Then $f_* : dH_*(X;R) \rightarrow dH_*(Y;R)$ is an $R$-module homomorphism as the Pontryagin algebras, and $f^* : dH^*(Y;R) \rightarrow dH^*(X;R)$ is an $R$-module homomorphism of coalgebras.
\end{thm}

\begin{proof} We need to show that $$f_*(r[x]) = rf_*([x]),$$ $$f_*([x_1] + [x_2]) = f_*([x_1]) + f_*([x_2])$$ and $$f_* ([x_1]\cdot_P [x_2]) = f_*([x_1])\cdot_P f_*([x_2])$$ for all $r$ in $R$ and $[x_1], [x_2]$ in $dH_*(X;R)$.
Since it is not difficult to show that the first and second conditions are satisfied, we will only check the final condition.
From the following commutative diagram
$$
\xymatrix@C=10mm @R=8mm{
dH_s(X;R) \otimes dH_s(X;R) \ar[d]_-{f_* \otimes f_*} \ar[r]^-{\mu_X} & dH_s(X \times X;R)  \ar[d]^-{(f\times f)_*} \ar[r]^-{{m_X}_*} & dH_s(X;R) \ar[d]^-{f_*} \\
dH_s(Y;R) \otimes dH_s(Y;R) \ar[r]^-{\mu_Y} & dH_s(Y \times Y;R) \ar[r]^-{{m_Y}_*} & dH_s(Y;R) \\
}
$$
of $R$-modules, we have
$$
\begin{array}{ll}
f_* ([x_1]\cdot_P [x_2])
               &= f_* ( {m_X}_* \circ \mu_X ([x_1] \otimes [x_2])) \\
               &= {m_Y}_* \circ \mu_Y \circ (f_* \otimes f_* )([x_1] \otimes [x_2]) \\
               &= {m_Y}_* \circ \mu_Y \circ (f_* ([x_1])\otimes f_* ([x_2]) \\
               &= f_*([x_1])\cdot_P f_*([x_2]): \\
\end{array}
$$
and similarly, for the coalgebra case.
\end{proof}

A {\it Hopf algebra} over $R$ is a graded $R$-algebra $A$ which has also the coalgebra structure whose $R$-algebra comultiplication $\varphi : A \rightarrow A \otimes A$ is a homomorphism of graded $R$-algebras. The graded Hopf algebras are often used in algebraic topology, and they have the natural algebraic structure on the direct sum of all homology or cohomology modules of a Hopf space.

\begin{rmk}
The relationship between the Pontryagin algebras and coalgebras of a digital Hopf space is that of dual Hopf algebras. Indeed, if $(Y,y_0,k_Y,m_Y)$ is a digital Hopf space, and $dH(Y;R)$ is a free graded $R$-module, then it can be shown that $dH_*(Y;R)$ and $dH^*(Y;R)$
are dual Hopf algebras.
\end{rmk}

\bigskip

\section{Conclusions}

Hopf spaces and co-Hopf spaces play a pivotal role in algebraic topology, especially in (equivariant) homotopy theory. Digital topology deals with the so-called discrete sets with a topology in the sense of Euclidean topology, which is considered to be digital sets or images of finite and bounded subsets of the $n$-dimensional Euclidean space. In digital topology, the digital process substitutes a bounded and finite discrete set for a suitable object in some category.

In the current study, we have developed the concept of a digital counterpart of classical notions in algebraic topology using the so-called algebraic approach from the classical homology, cohomology and Pontryagin algebra. More specifically, This study focused on setting up of more algebraic invariants and their fundamental properties of digital homology and cohomology modules over a commutative ring with identity for a digital image with an adjacent relation which are based on the classical homology and cohomology groups of topological spaces in algebraic topology.

% The following MDPI journals use author-date citation: Arts, Econometrics, Economies, Genealogy, Humanities, IJFS, JRFM, Laws, Religions, Risks, Social Sciences. For those journals, please follow the formatting guidelines on http://www.mdpi.com/authors/references
% To cite two works by the same author: \citeauthor{ref-journal-1a} (\citeyear{ref-journal-1a}, \citeyear{ref-journal-1b}). This produces: Whittaker (1967, 1975)
% To cite two works by the same author with specific pages: \citeauthor{ref-journal-3a} (\citeyear{ref-journal-3a}, p. 328; \citeyear{ref-journal-3b}, p.475). This produces: Wong (1999, p. 328; 2000, p. 475)

%%%%%%%%%%%%%%%%%%%%%%%%%%%%%%%%%%%%%%%%%%
%% optional
%\sampleavailability{Samples of the compounds ...... are available from the authors.}

%% for journal Sci
%\reviewreports{\\
%Reviewer 1 comments and authors’ response\\
%Reviewer 2 comments and authors’ response\\
%Reviewer 3 comments and authors’ response
%}

%%%%%%%%%%%%%%%%%%%%%%%%%%%%%%%%%%%%%%%%%%
\end{document}